\documentclass[9pt]{extarticle}
\usepackage{spconf,amsmath,graphicx,etoolbox,placeins,float}
\apptocmd{\thebibliography}{\setlength{\itemsep}
{2pt}\setlength{\parskip}{0pt}}{}{}
\usepackage{stfloats}
\usepackage{caption}
\usepackage{multirow}
\usepackage{adjustbox}
\usepackage{hyperref}   

\title{A Benchmark for Joint Dialogue Satisfaction, Emotion Recognition, and Emotion State Transition Prediction}
\name{%
\begin{tabular}{c}
Jing Bian$^{1,3}$\sthanks{Equal Contribution}, 
Haoxiang Su$^{1,2,3\ast}$\sthanks{Internship at China Telecom Corp Ltd.}, 
Liting Jiang$^{1,3}$, Di Wu$^{1,2,3\dagger}$, Ruiyu Fang$^{2}$, Xiaomeng Huang$^{2}$, \\
Yanbing Li$^{1,3}$, Shuangyong Song$^{2,\ddagger}$, Hao Huang$^{1,3,4}$\sthanks{Corresponding authors. Supported by NSFC (62466055) and State Language Commission (ZDI145-96).}%
\end{tabular}%
}
\address{
\textsuperscript{1}School of Computer Science and Technology, Xinjiang University, Urumqi, China \\
\textsuperscript{2}Institute of Artificial Intelligence (TeleAI), China Telecom Corp Ltd, Financial Street, Beijing, China\\
\textsuperscript{3}Joint International Research Laboratory of Silk Road Multilingual Cognitive Computing, Urumqi, China\\
\textsuperscript{4}Xinjiang Key Laboratory of Multi-lingual Information Technology, China
}

\begin{document}
%
\maketitle

\begin{abstract}
    
User satisfaction is closely related to enterprises, as it not only directly reflects users’ subjective evaluation of service quality or products, but also affects customer loyalty and long-term business revenue. Monitoring and understanding user emotions during interactions helps predict and improve satisfaction. However, relevant Chinese datasets are limited, and user emotions are dynamic; relying on single-turn dialogue cannot fully track emotional changes over multiple turns, which may affect satisfaction prediction. To address this, we constructed a multi-task, multi-label Chinese dialogue dataset that supports satisfaction recognition, as well as emotion recognition and emotional state transition prediction, providing new resources for studying emotion and satisfaction in dialogue systems.\footnote{\url{https://github.com/EST-DATASET/EST-DATASET}}.

\end{abstract}
\begin{keywords}
User satisfaction, emotion recognition, emotional state transformations, large language model
\end{keywords}
\section{Introduction}
\label{sec:intro}
In task-oriented dialogue systems, dialogue state tracking\cite{song2024improving,song2024graph,su2023scalable} and user intent recognition\cite{wu2025int,xie2025mitigating} are core tasks for understanding user needs and supporting system decisions and interaction flows, while user satisfaction is an important metric for measuring users’ subjective evaluation of products or services and plays a crucial role in customer service and dialogue systems\cite{shankar2003customer,yao2020session}. Some studies focus on user intent\cite{guo2020deep,song2017intension} or sequence modeling of user behaviors\cite{deng2022user}. With the development of large language models (LLMs)\cite{he2024telechat,wang2024telechat,liu2025training}, using LLMs for user satisfaction prediction has become a new research direction\cite{abolghasemi2024cause,lin2024interpretable}. For example, \cite{abolghasemi2024cause}leveraged LLMs to augment data by generating counterfactual samples to balance the distribution of satisfied and dissatisfied samples; \cite{lin2024interpretable} achieved more accurate and interpretable satisfaction assessment.

\begin{figure}[t]
  \centering
  \includegraphics[scale=1, trim=0pt 0pt 0pt 0pt, clip]{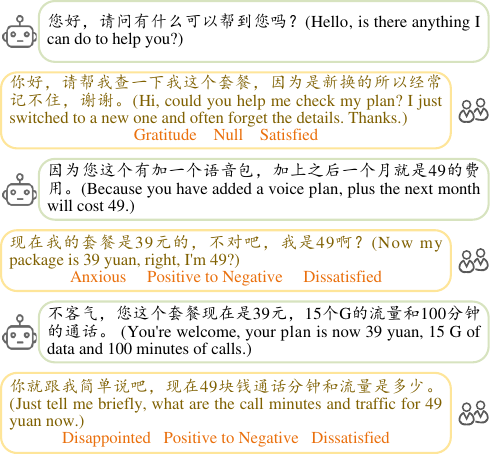}
  \caption{An example from the dataset. The green box represents the customer service representative, and the yellow box represents the user. The user's initial emotional state transition is empty. We annotate each user's utterance with emotion, emotional state transition, and satisfaction.}
  \label{fig:framework1}
\end{figure}

Although these works represent key advances in dialogue semantic understanding, satisfaction prediction over long-term conversations remains challenging without access to user-level emotional information. Introducing user emotions\cite{jiang2025label,jiang2025utterance} as auxiliary information can provide effective cues for prediction\cite{song2023speaker}. For instance,\cite{song2023speaker} regards satisfaction assessment and sentiment analysis as a collaborative task and proposes a multi-task adversarial strategy.

However, existing Chinese dialogue datasets that capture both user satisfaction and emotions are limited. In real-world interactions, user emotions are dynamic, and single-turn dialogues are insufficient to fully capture emotional changes across multiple turns. Accessing users’ emotional dynamics can help track these changes more accurately, enabling customer service agents to provide appropriate responses, ease user emotions, and improve the accuracy of satisfaction prediction.

To address this issue, we simulated interactions between real users and customer service representatives to create a multi-task Chinese dialogue dataset. The dataset covers five common service categories: business inquiries, service processing, complaint and suggestion handling, fault reporting and technical support, and customer care and follow-up. It includes annotations for user emotions in each turn, emotional state transitions, and user satisfaction, supporting three tasks: emotion recognition, emotional state transition prediction, and satisfaction prediction. With this dataset, researchers can systematically analyze the relationship between user emotions and satisfaction, providing data support for modeling user experience, emotion prediction, and optimizing intelligent customer service systems in multi-turn dialogues. To our knowledge, this is the first Chinese dialogue dataset annotated with user emotional state transitions.

\renewcommand{\arraystretch}{1.1}
\begin{table*}
\centering
\scriptsize
\caption{\centering Comparison with existing datasets.The table presents datasets commonly used for emotion recognition and satisfaction prediction.}
\label{tab:topic1}
\resizebox{1\textwidth}{!}{%
\begin{tabular}{l|c|c|c|c|c|c|c}
\hline
\textbf{Dataset} & \textbf{Modalities} & \textbf{Type} & \textbf{Multi-label}& \textbf{language}& \textbf{Total Dialogue}& \textbf{Annotate Utterances} &\textbf{Turn}\\
\hline
EmoryNLP\cite{zahiri2018emotion} & Text & Chit-chat & Yes & English & \textbackslash & 12,606 & \textbackslash\\
EmotionLines\cite{hsu2018emotionlines} & Text & Chit-chat & No & English & 2,000 & 29,245& \textbackslash\\
DailyDialog\cite{li2017dailydialog} & Text & Chit-chat & No & English & 13,118 & 102,979 & 103,632  \\
EmoWOZ\cite{feng2022emowoz}& Text & Task-oriented & No & English & 11,434 & 83,617 & 167,234  \\
MELD\cite{poria2019meld} & Audio,Visual,Text & Chit-chat & No & English & 1,433 & 13,708  & \textbackslash\\
IEMOCAP\cite{busso2008iemocap}& Audio,Visual,Text & Chit-chat & No & English & \textbackslash & 7433 &\textbackslash \\
CH-SIMS\cite{yu2020ch}& Audio,Visual,Text & Chit-chat & No & English & \textbackslash & 2,281 &\textbackslash \\
M\textsuperscript{3}ED\cite{zhao2022m3ed} & Audio,Visual,Text & Chit-chat & Yes & Chinese & 990 & 24,449 & 9082 \\
\hline
\hline
Clothes\cite{song2019using} & Text & Task-oriented & Yes & Chinese  & 10,000 & \textbackslash  & \textbackslash \\
Makeup\cite{song2019using}  & Text & Task-oriented & Yes & Chinese  & 3,540 & \textbackslash & \textbackslash\\
USS\cite{sun2021simulating}& Text & Task-oriented & No & E and C & 6,800 & \textbackslash   & 99,569 \\
\hline
\textbf{Ours} & Text & Task-oriented & Yes & Chinese  & 90,000 & 1,590,895 & 1,240,327  \\
\hline
\end{tabular}
}    
\end{table*}

\renewcommand{\arraystretch}{1.1}
\begin{table}[H]
\centering
\caption{Basic statistics of dataset}
\label{tab:2}
\resizebox{0.3\textwidth}{!}{%
\begin{tabular}{l|c}
\hline
\textbf{Metrics} & \textbf{Dataset} \\
\hline
Total sessions & 90,000 \\
Total turns & 1,240,327 \\
Total user utterances & 1,590,895 \\
Average turns per session & 13.78 \\
\hline
\end{tabular}
}
\end{table}

\section{RELATED WORK}
Table~\ref{tab:topic1} summarizes information about datasets commonly used in emotion recognition and user satisfaction research. The section above the double line lists sentiment analysis datasets, including four multi-modal datasets and four text-based datasets. It can be observed that, whether multi-modal or text-based, these datasets primarily consist of casual conversational data, lacking representation of structured dialogue features. 

The section below the double line lists user satisfaction datasets. The USS dataset extracts portions of data from five commonly used task-oriented dialogue datasets—JDDC, SGD, ReDial, MWOZ, and CCPE—and then annotates user satisfaction. During dataset construction, a labeled classifier is used to identify user emotions, dialogues that do not contain negative sentiment are filtered out, and the remaining dialogue content is manually annotated. The Clothes and Makeup datasets are both collected from leading e-commerce platforms in China. Overall, the majority of these datasets are in English, with relatively few available in Chinese.

\section{DATASET CONSTRUCTION}
\label{sec:format}

\subsection{Data Collection}

We constructed a dataset covering typical customer service issues from telecom operators. The data was annotated over a five-month period. A complete dataset was synthesized by transcribing call recordings between telecom customer service representatives and users using speech recognition and emulating their interactions. The data covers five common service categories: business consultation, business processing, complaint and suggestion handling, fault reporting and technical support, and customer care and return visits. In actual conversation scenarios, the opening of a conversation is not fixed. Customer service representatives do not always initiate the conversation with a greeting; it is also common for users to directly ask questions as the starting point of the conversation. Therefore, in this dataset, each complete conversation is preserved in the form of natural interaction.

As shown in Table~\ref{tab:2} the dataset collected in this study contains 90,000 complete conversations, totaling 1,240,327 turns, with an average of 13.78 turns per conversation, and a total of 1,590,895 user utterances were annotated. Although the data were synthesized to emulate real user interactions, it may still contain information such as phone numbers that appear in real life. Therefore, all conversations were rigorously desensitized. For example, customer phone numbers and ID numbers appearing in the dataset have been replaced with the special token \texttt{[num]}.

\subsection{Annotate Details And Consistency}
The annotation process is divided into three stages: emotion annotation, emotion state transition annotation, and satisfaction annotation. We initially designed three schemes for the classification of emotion categories. The first scheme adopts the six basic emotions proposed by American psychologist Paul Ekman, namely \textit{Anger},\textit{ Disgust},\textit{ Fear}, \textit{Happiness}, \textit{Sadness}, and \textit{Surprise}. The second and third schemes refer to the existing emotion recognition datasets  MELD\cite{poria2019meld} and IEMOCAP\cite{busso2008iemocap} . Among them, IEMOCAP further adds four categories of emotions: \textit{Frustration}, \textit{Excited},\textit{Neutral state} and \textit{Others} on the basis of the six basic emotions; MELD introduces one category of emotions: \textit{Neutral}. However, the classification methods of the above two datasets are not fully applicable to the dialogue dataset we collected. In actual scenarios, users usually call customer service to seek solutions to their problems, so the frequency of emotions such as \textit{Surprise} and \textit{Expectation} is low; at the same time, because customer service answers may not meet user needs, abusive language expressions also appear in the corpus. Since the satisfaction labels include a \textit{Neutral} category, we use \textit{No Emotion} instead of \textit{Neutral} in the emotion labels to avoid ambiguity and ensure clear distinction between the two label sets. Based on this, combined with the contextual characteristics of the conversation between customer service and users, the fine-grained emotion categories were finally divided into seven categories: \textit{Worry}, \textit{Anger}, \textit{Insult}, \textit{Disappointment}, \textit{Anxiety}, \textit{Gratitude} and \textit{No Emotion}, and each round of user discourse was labeled accordingly.

\begin{figure*}[t]
  \centering
  \includegraphics[width=\textwidth]{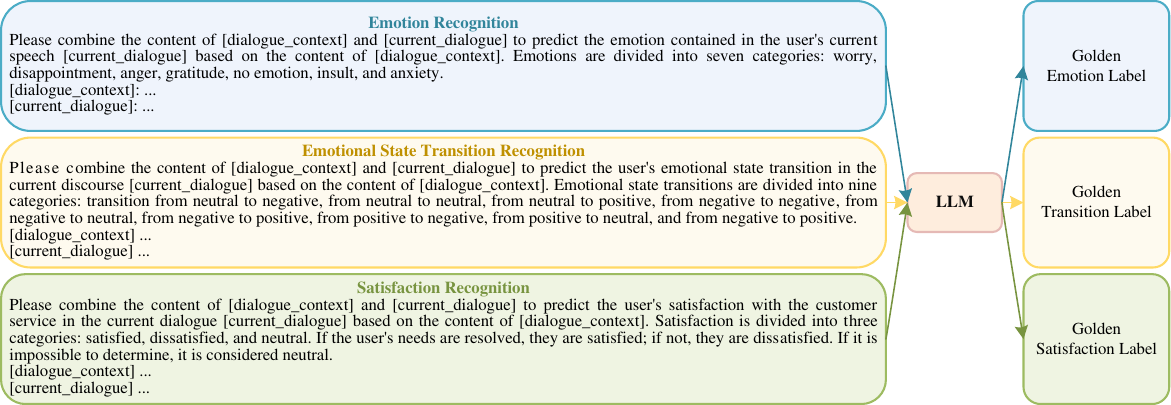}
  \caption{Model architecture diagram. Blue boxes represent the emotion recognition task, yellow boxes represent the emotion state transition task, and green boxes represent the satisfaction prediction task.
}
  \label{fig:framework2}
\end{figure*}

After completing fine-grained sentiment annotation of user utterances, we discovered a significant imbalance in their sentiment distribution, with the majority of user utterances labeled as \textit{neutral}. To improve the effectiveness and generalization of the annotation system, we further mapped fine-grained sentiment categories into three sentiment polarities: \textit{Positive (Gratitude)}, \textit{Neutral (No Emotion)}, and \textit{Negative} (\textit{Worry}, \textit{Anger}, \textit{Insult}, \textit{Disappointment}, and \textit{Anxiety)}. In the second phase of sentiment state transition annotation, based on these three sentiment polarities, sentiment transitions can be divided into nine types. The specific label types are shown in Table \ref{tab:topic3}. To ensure the rationality of sentiment transition annotation, special processing is applied to the initial turn of the conversation: if the user utterance in turn 0 contains only polite expressions or short sentences without substantive meaning (such as “\textit{hello}” or “\textit{hmm}”), subsequent turns are observed until information with clear semantics appears, and this is used as the initial sentiment polarity of the conversation. Therefore, sentiment transition refers to the change in the emotional state expressed by the user's current utterance compared to the initial sentiment-laden utterance in the conversation.

Satisfaction marking in the third stage. Satisfaction prediction is closely related to sentiment analysis. The sentiment expressed in user utterances is an important basis for judging their satisfaction. Therefore, referring to the method of\cite{song2019using}, fine-grained sentiment is further mapped to satisfaction polarity: when the fine-grained sentiment is \textit{Gratitude}, it is annotated as \textit{Satisfied}; when the sentiment is \textit{No Emotion}, it is annotated as \textit{Neutral}; when the sentiment is \textit{Worry}, \textit{Anger}, \textit{Insult},\textit{ Disappointment} or \textit{Anxiety}, it is annotated as \textit{Dissatisfied}. Overall, the dataset contains 1,590,895 user utterances, of which the proportion of fine-grained sentiment being \textit{No Emotion} is about 96.3\%, and the proportion of sentiment states changing from \textit{Neutral to Neutral }is about 80\%. This distribution feature is highly consistent with real-world application scenarios, because most users' interactions with customer service are mainly focused on consulting business-related issues rather than emotional expressions. The number of annotations for each task label in the dataset and the data used in the experiment are shown in Table \ref{tab:topic3}.

Fine-grained sentiment labeling and emotion state transition labeling are completed by an outsourced team. To ensure labeling quality, labelers are required to conduct internal cross-verification after completing their respective tasks. For samples that remain questionable after cross-verification, multiple senior labeling experts will be invited to conduct a review to finalize the labeling results. In the satisfaction labeling phase after data screening, different fine-grained sentiments are first automatically mapped to corresponding satisfaction polarities through a script. Two graduate students in computer science majors then manually review the results and correct any errors found. This process is carried out in an iterative manner to effectively ensure the accuracy of the satisfaction labels.

\begin{table}[t]
\centering
\caption{\textbf{Experimental data label distribution}}
\label{tab:topic3}
\resizebox{\linewidth}{!}{%
\begin{tabular}{l|c|c|c}
\hline
\textbf{Tasks} & \textbf{Types} & \textbf{Annotation} & \textbf{Used in experiment} \\
\hline
\multirow{7}{*}{Emotion} 
& Worry & 30,486 & 13,696 \\
& Anger & 2,918 & 2,911 \\
& Insult & 275 & 275 \\
& Disappointed & 10,096 & 10,017 \\
& Anxiety & 1,898 & 1,870 \\
& Gratitude & 11,661 & 11,561 \\
& No Emotion& 1,533,561 & 10,892 \\
\hline
\multirow{9}{*}{Emotion transfer} 
& Positive to Neutral & 1,589 & 265 \\
& Positive to Positive & 162 & 42 \\
& Positive to Negative & 54 & 43 \\
& Neutral to Neutral & 1,189,094 & 9,914 \\
& Neutral to Positive & 45,552 & 8,883 \\
& Neutral to Negative & 43,106 & 18,603 \\
& Negative to Neutral & 201,209 & 975 \\
& Negative to Positive & 7,326 & 1,605 \\
& Negative to Negative & 12,735 & 10,892 \\ 
\hline
\multirow{3}{*}{Satisfaction} 
& Satisfied & 11,661 & 11,561 \\
& Dissatisfied & 45,673 & 28,769 \\
& Neutral & 1,533,561 & 10,892 \\
\hline
\end{tabular}%
}
\vspace{0.45cm}
\end{table}
\FloatBarrier

\begin{table*}[!t]
\centering
\small 
\caption{\centering Main experiment results. Experimental results of eight LLMs and two models related to satisfaction}
\label{tab:4}
\resizebox{\textwidth}{!}{%
\begin{tabular}{c|c|cccc|cccc|cccc}
\hline
\textbf{Model} &  & \multicolumn{4}{c|}{\textbf{Emotion}} & \multicolumn{4}{c|}{\textbf{Emotion Transfer}} & \multicolumn{4}{c}{\textbf{Satisfaction}} \\
\hline
 & \textbf{$\textbf{ALL\_A}$} &\textbf{ A} & \textbf{P }&\textbf{ R} & \textbf{F1} & \textbf{A} & \textbf{P} & \textbf{R} & \textbf{F1} & \textbf{A} & \textbf{P} & \textbf{R} & \textbf{F1} \\
\hline
ASAP\cite{ye2023modeling} & \textbackslash & 0.6074 & 0.5894 & 0.6297 & 0.5429 & 0.5822 & 0.5822& 0.5822& 0.5822 & 0.7267& 0.8288 & 0.7329 & 0.6842\\
USDA\cite{deng2022user} & \textbackslash & 0.6014 & 0.6014 & 0.6014 &  0.6014 & 0.5719 & 0.3693 & 0.4566 & 0.3867 & 0.7647 & 0.8283 & 0.7364 & 0.6863 \\
\hline
\hline
Baichuan2-7B\cite{yang2023baichuan} & 0.6744 & 0.5695 & 0.5340 & 0.5244 & 0.4646 & 0.6360 & 0.7284 & 0.5105 & 0.5044 & 0.7350 & 0.8082 & 0.7058 & 0.6601  \\
GLM4-9B\cite{glm2024chatglm} & 0.6573 & 0.5776 & 0.5507 & 0.5505 & 0.4879 & 0.6538 & 0.7638 & 0.5568 & 0.5620 & 0.7404 & 0.8110 & 0.7115 & 0.6677 \\
Deepseek\cite{guo2025deepseek} & 0.6384 & 0.5609 & 0.4634 & 0.4447 & 0.3970 & 0.6110 & 0.4249 & 0.3553 & 0.3304 & 0.7301 & 0.8065 & 0.6965 & 0.6503 \\
Mistral-7B\cite{jiang2023mistral7b} & 0.6643 & 0.6084 & 0.5516 & 0.5817 & 0.5040 & 0.6215 & 0.7576 & 0.5369 & 0.5505 & 0.7634 & 0.8280 & 0.7353 & 0.6863 \\
TeleChat2-7B\cite{wang2025technical} & 0.6686 & 0.6126 & 0.5532 & 0.5926 & 0.5132 & 0.6415 & 0.7611 & 0.5385 & 0.5621 & 0.7655 & 0.8367 & 0.7457 & 0.6883 \\
Qwen-7B\cite{bai2025qwen2} & 0.6483 & 0.5644 & 0.5508 & 0.5127 & 0.4612 & 0.6457 & 0.5635 & 0.4407 & 0.4323 & 0.7333 & 0.8052 & 0.7042 & 0.6612 \\
LLaMa2-7B\cite{touvron2023llama} & 0.7208 & 0.5743 & 0.5039 & 0.5000 & 0.4402 & 0.7426 & 0.6614 & 0.4653 & 0.4576 & 0.8455 & 0.8766 & 0.8273 & 0.8183 \\
LLaMa3-8B\cite{dubey2024llama} & 0.6624 & 0.6106 & 0.5813 & 0.6069 & 0.5336 & 0.6658 & 0.7325 & 0.5859 & 0.5844 & 0.7647 & 0.6213 & 0.5523 & 0.5148 \\
\hline
\end{tabular}
}
\end{table*}

\section{Experimental}
\subsection{Experimental Setup}
The development of LLMs enables models to handle multiple tasks within a single framework, no longer limited to a single function or objective. This multi-task capability not only facilitates the understanding of complex dialogues but also significantly enhances the model’s versatility and efficiency. Using this advantage, as shown in Table~\ref{tab:topic3}, the dataset was divided into training, validation, and test sets with a ratio of 8:1:1.The model was fine-tuned using the LoRA approach on two NVIDIA RTX 3090 GPUs, with the rank set to 4. Optimization was carried out with a learning rate of $10^{-5}$ and a dropout rate of 0.1.The batch size per device was set to 1, and with a gradient accumulation step of 4,the effective batch size resulting was 4.For training monitoring and model recovery, logging was carried out every 5 steps,and model checkpoints were saved every 100 steps.

Although the prediction task is inherently a classification problem, this study employed a LLM to accomplish it in a generative manner. Figure \ref{fig:framework2} is the model architecture diagram. Specifically, the experiment used prompts to clearly define candidate categories for each task and required the LLM to generate one of these categories for each input sentence. This approach, through prompt engineering, enabled the generative model to adapt to classification tasks\cite{xiong2024dual}. The experimental evaluation process used commonly used evaluation metrics for classification tasks, including accuracy (A), precision (P), recall (R), and the macro-F1 score (F1), to ensure the scientific nature and comparability of the results.

\subsection{Baselines}

In this study, we selected eight LLMs and two two embedding-based models related to user satisfaction (ASAP and USDA) for validation and comparative analysis. The specific model selection is shown in Table~\ref{tab:4}. The ASAP model and the USDA model are commonly used models for the satisfaction task. The former is an embedding-based method that incorporates a Hawkes Process in modeling user satisfaction. The latter is also based on an embedding model and jointly optimizes the sequential dynamics of user satisfaction and user behavior.

\subsection{Experimental Analysis}

The experimental results are summarized in Table~\ref{tab:4}. Key observations are as follows:

\begin{enumerate}
    \item \textbf{Task adaptation:} 
   To enhance the usability of the dataset, we selected eight widely used LLMs and two deep learning models related to user satisfaction (ASAP and USDA) for validation and comparative analysis. ASAP and USDA were originally designed for analyzing user satisfaction and user intent. In this experiment, the original user intent recognition task of these two pre-trained models was replaced with an emotion recognition task, and an emotional state transition task was added, while all other training parameters remained unchanged.

    \item \textbf{Overall performance:} 
    LLaMa2 achieved the highest overall accuracy and performed most effectively on the satisfaction task, with a macro-F1 score of 0.8183, outperforming all other models. LLaMa3 achieved the best performance on the emotion and emotion transfer tasks, with macro-F1 scores of 0.5336 and 0.5844, respectively, demonstrating strong capability in fine-grained emotion and state transition modeling. Other LLMs performed moderately overall but were competitive on individual tasks. Although ASAP and USDA performed weaker than the LLMs overall, their macro-F1 scores on the satisfaction task were close to those of the LLMs.

    \item \textbf{Task-specific difficulty:} 
    Among the three tasks, the emotion state transfer task was the most challenging, with all models achieving generally low macro-F1 scores. The satisfaction prediction task was relatively easy, while the emotion recognition task fell in between, showing larger performance differences across models.

    \item \textbf{Challenges and future directions:} 
    Despite their strong performance, LLMs still face several challenges. Experimental results indicate that modeling emotional dynamics in multi-turn conversations remains difficult, especially for the sentiment transfer task. Additionally, the dataset exhibits label imbalance across categories, which may affect model robustness. Future work could explore strategies such as multi-task learning, parameter sharing, and data imbalance mitigation to further improve model performance and leverage strengths across tasks.
\end{enumerate}

\section{Conclusion}
\label{sec:majhead}

User satisfaction is a key metric for evaluating users’ subjective experience, and incorporating emotional information can improve satisfaction prediction. However, existing Chinese datasets are limited and fail to capture the dynamic nature of emotions in real conversations. In this work, we construct a multi-task Chinese emotional dialogue dataset that supports emotion recognition, emotion state transition recognition, and user satisfaction prediction, and is the first to annotate emotion state transitions. We further conduct multi-task experiments based on large language models to evaluate model performance and explore future research directions.

\clearpage


\bibliographystyle{IEEEtran}

\bibliography{main}

\end{document}